\documentclass{article}
\usepackage{arxiv_template}
\BiblatexSplitbibDefernumbersWarningOff

\title{%
On Distinguishing Capability Elicitation from Capability Creation \\ in Post-Training: A Free-Energy Perspective
}

\author{
\name{Yuhao Li} \and \name{Shengchao Liu}\\
\addr{Department of Computer Science and Engineering} \\
\addr{The Chinese University of Hong Kong} \\
\email{\{yuhao.li, scliu\}@cuhk.edu.hk}
}

\begin{document}

\maketitle

\begin{abstract}
Debates about large language model post-training often treat supervised fine-tuning (SFT) as imitation and reinforcement learning (RL) as discovery. But this distinction is too coarse. What matters is whether a training procedure increases the probability of behaviors the pretrained model could already produce, or whether it changes what the model can practically reach. \textbf{We argue that post-training research should distinguish between capability elicitation and capability creation.}
We make this distinction operational by introducing the notion of accessible support: the set of behaviors that a model can practically produce under finite budgets. Post-training that reweights behaviors within this support is capability elicitation; whereas changing the support itself corresponds to capability creation. 
We develop this argument through a free-energy view of post-training. 
SFT and RL can both be seen as reweighting a pretrained reference distribution, only with different external signals. Demonstration signals define low-energy behavior for SFT, and reward signals define low-energy behavior for RL. 
When the update remains close to the base model, the main effect is local reweighting, not capability creation. Within this framework, the central question is no longer whether post-training is framed as SFT or RL, but whether it reweights behaviors already within reach, or instead expands the model's reachable behavioral space through search, interaction, tool use, or the incorporation of new information.
\end{abstract}

\section{Introduction}

Post-training has become a central stage in building large language models (LLMs). After large-scale pretraining, LLMs are commonly further shaped through supervised fine-tuning \citep{GPT3, Llama2}, preference optimization \citep{DPO, ziegler2020finetuning}, or reinforcement learning from human feedback \citep{Stiennon2020Learning, InstructGPT}. 
These procedures can substantially change model behavior, sometimes producing large gains in instruction following \citep{wei2022finetuned, Self-Instruct}, coding \citep{codex, CodeLlama}, and mathematical problem solving \citep{lightman2024verify, DeepSeekMath}.
This raises a basic question about what post-training actually does. Does it elicit behaviors that were already present but unreliable in the pretrained base model, or create the capabilities that the base model could not previously reach?

A common way to frame this question is through the contrast between supervised fine-tuning (SFT) and reinforcement learning (RL) \citep{mayi2025sft-rl, jiang2026supervised}. 
Both of these methods are applied after pretraining, further shaping the base model's behaviors.
SFT is often described as imitation from demonstrations: the model is trained to reproduce behaviors present in a dataset \citep{LIMA, Zhang2026Instruction}. RL is often associated with reward-guided exploration and discovery \citep{wang2025reinforcement}. Candidate behaviors can be sampled or searched for, and then reinforced if they receive a high reward.
This contrast is especially salient in recent discussions of reasoning models \citep{OpenAI-o1, DeepSeek-R1}, where RL-style training is often credited with producing stronger reasoning behavior.
It is therefore tempting to treat SFT as a mechanism for eliciting existing capabilities and RL as a mechanism for creating new ones.

However, we argue that this identification is too coarse. Demonstrations and rewards describe the form of the training signal, but they do not determine the capability mechanism. 
The more relevant object is the set of behaviors that a model can practically produce under finite sampling, optimization, and divergence budgets. We introduce the concept of \emph{accessible support} to formalize this idea. 
If post-training only increases the reliability of behaviors in the accessible support, we define it as \emph{capability elicitation}. Instead, if it changes the accessible support itself, making previously unreachable behaviors accessible, we define it as \emph{capability creation}.

Under this criterion, SFT and RL are algorithmic categories, whereas elicitation and creation are claims about how post-training changes the model's behavioral distribution. 
A supervised loss in SFT can also fit human-written answers, verifier-filtered samples, search-generated solutions, tool-assisted trajectories, or data distilled from stronger models \citep{STaR, sharma2025eliciting, OpenMathInstruct2, ToRA}. Conversely, a reward objective in RL can also only make existing behaviors more likely. 
Thus, a performance gain after SFT or RL does not by itself determine whether a new capability has been created. The performance gain depends on whether or not the evaluation metric can reflect the corresponding change over the behavioral distribution (elicitation or creation), regardless of the algorithmic categories (SFT or RL).

\textbf{Our position: Post-training research should distinguish capability elicitation from capability creation: does a method reweight already accessible behaviors or expand what the model can reach?}
In this view, the central question is not whether a method is called SFT or RL. Rather, it is what training signal the task provides, where its candidate behaviors are drawn from, and whether the post-training pipeline expands the accessible behavioral space of the base model. 

We formalize this position from a free-energy perspective on post-training.  Related variational forms have already been established in maximum-entropy RL \citep{Theodorou2012Relative, Maximum-entropy-RL} and KL-regularized policy optimization \citep{DPO}. 
Our use of this perspective is diagnostic. It clarifies that demonstration fitting and reward optimization can both be understood as reshaping the pretrained distribution under an external training signal. Demonstrations define low-energy behavior for SFT, while rewards define low-energy behavior for RL. When the update remains close to the reference model, the primary effect is local reweighting rather than capability creation.

The rest of the paper develops this position as follows. 
We first clarify why the SFT--RL debate confounds algorithmic form with capability mechanism: SFT and RL specify different training signals, but neither label alone determines whether a gain reflects elicitation or creation. 
Then we introduce a free-energy perspective that describes SFT and KL-regularized RL as local reweighting of a pretrained reference distribution. 
Building on this view, we use accessible support to distinguish different kinds of reachability, leading to four regimes of post-training: demonstration-covered elicitation, tail reweighting, barrier-crossing discovery, and unsupported regimes. 
Finally, we discuss alternative views and clarify the scope of the framework.

\section{The Capability Debate In Post-Training}

The SFT--RL contrast is useful, but it does not by itself explain why post-training improves a model \citep{ethayarajh2024model, Gheshlaghi2024General}. SFT and RL specify different optimization objectives, yet a post-training pipeline contains more than the objective. It also contains the source of data, the procedure for generating candidate behaviors, and the signal for selecting among them \citep{gudibande2024false, ryd2026removing}. The more important question is therefore not whether a method is called SFT or RL, but what kind of information the post-training provides \citep{korbak2022rl-kl}. 

\textbf{Notation.} Before analyzing these signals, we establish standard notations. Let $x$ denote an input prompt and $y$ denote the response. We use $p_0(y|x)$ to represent the conditional distribution of the pretrained base model, which also serves as a reference model in RL. The behavioral distribution after post-training is denoted as $q(y|x)$, parameterized as $p_\theta(y|x)$ in practice.

\subsection{The Capability of SFT Depends on the Demonstration Distribution}

Supervised fine-tuning usually minimizes the negative log-likelihood on demonstration data:
\begin{equation} \label{eq:sft_loss}
\mathcal{L}_{\mathrm{SFT}}(\theta) = -\mathbb{E}_{p_{\rm demo}(x,y)}
\log p_\theta(y|x),
\end{equation}
where $p_{\rm demo}(x,y)$ is the empirical demonstration distribution.
From the form of the loss, SFT is indeed imitating demonstrations. But imitation is not synonymous with weakness. The supervised objective does not specify where the target behaviors come from. It only specifies that the model should increase the target behaviors' likelihood once they appear in the demonstration distribution.

This makes $p_{\rm demo}$ the central object. Different construction procedures induce different demonstration distributions, with different information content. 
If the demonstrations contain trajectories generated with richer or more informative methods, the same loss function can elicit behaviors that the base model rarely or never produces.

Thus, the limitation of SFT arises not from the word \emph{supervised} itself, but from the inability of the demonstration distribution to adequately cover all target behaviors, especially the low-probability ones.
\begin{itemize}[leftmargin=*, noitemsep, topsep=0pt]
    \item If the demonstrations densely cover the desired behavior, SFT can efficiently elicit and stabilize it.
    \item If the demonstrations carry additional or more complex information than the base dataset, SFT may trigger behaviors that the original training set would not produce.
    \item However, if they contain no positive evidence for the target behavior, the supervised loss supplies no discovery signal by itself.
\end{itemize}

This means that any comparison between SFT and RL is ill-defined if it does not specify the source of demonstrations. When discussing the capability limits of SFT, the key question is not whether it imitates, but what distribution it imitates and how that distribution is constructed.

\subsection{The Capability of RL Is Constrained by the Reference Distribution}

Reinforcement learning is often assigned the opposite meaning: exploration and discovery. This interpretation is reasonable when an agent truly interacts with an environment, generates new trajectories, and receives external feedback. But many forms of language-model post-training are not unconstrained exploration. They usually optimize rewards or preferences near a reference model \citep{ziegler2020finetuning, DPO}
\begin{equation} \label{eq:rl_loss}
\max_q \;
\mathbb{E}_{y\sim q(y|x)}[R(x,y)]
- \beta \, \mathrm{KL} \left(q(y|x) \,\|\, p_0(y|x)\right).
\end{equation}
Here, $R(x,y)$ is a reward score assigned to the response $y$, and $\beta>0$ controls the strength of the KL constraint, thereby limiting how far the optimized distribution $q(y|x)$ can deviate from the reference model $p_0(y|x)$.
The key issue is not whether \emph{reward is used}, but how much freedom reward optimization has relative to the reference distribution.

When the KL constraint is strong, RL does not freely search over all possible responses. It mainly reweights responses within the behavior class already induced by the base model. 
A response that is initially rare but high-reward can therefore become frequent after training. Such a change may produce a large benchmark gain, but it only reweights already accessible behaviors rather than enabling new behaviors that were previously unreachable.

In contrast, if the training pipeline introduces exploration mechanisms that sample trajectories outside the high-probability manifold of $p_0 (y|x)$, the resulting gain reflects capability creation. Because the reachable candidate set has expanded, the optimization is no longer bound by local distribution reweighting.

This distinction clarifies why not all uses of RL should be interpreted in the same way. Although both procedures may be called RL-based post-training, they do not make the same claim about capability.

\subsection{The Hidden Axis: Elicitation Versus Creation}

Therefore, the more important axis in the post-training debate is not supervision versus reinforcement, but capability elicitation versus capability creation, {\ie{}}, whether post-training primarily increases the likelihood of already accessible behaviors, or expands the set of behaviors the model can reach.

This axis cuts across the existing mainstream SFT/RL distinction for performance \citep{mayi2025sft-rl, jiang2026supervised}. 
SFT is not automatically mere elicitation: if its demonstrations are of sufficiently high quality, the supervised objective may transmit behaviors that the target model would not have produced on its own \citep{sharma2024critical, LIMO}.
Conversely, RL is not automatic discovery either: under strong reference constraints, reward optimization may only amplify behaviors already available for the base model \citep{korbak2022rl-kl}.

The next two sections develop the diagnostic framework needed to distinguish these cases: first through a free-energy perspective on local reweighting, and then through accessible support.

\section{The Free-Energy Perspective}

To formally distinguish capability elicitation from creation, we need a mathematical language to describe how post-training modifies the behavioral distribution of the base model. Here, we adopt a free-energy perspective \citep{Mazzaglia2022, taniguchi2025generative, elliott2026stagewise}, which allows us to interpret many post-training methods as reweighting a pretrained reference distribution. 

In this perspective, a policy does not sample arbitrarily from all possible behaviors. Rather, it redistributes probability mass according to reward or energy relative to a reference distribution or prior dynamics.
Our novel utilization of this framework is to diagnose the capability claims in post-training.
When a model performs better after SFT, DPO, RLHF, or other post-training methods, we first ask whether the improvement can be explained as a reweighting of a reference distribution.

\subsection{Preliminary for Free-Energy Framework}

Many post-training objectives can be understood in analogy to the physical concept of free energy, commonly written as $F = E - TS$, where $E$ is the energy of system, $S$ is the entropy, and $T$ is the temperature parameter. 

Adapting this to the post-training, $E$ corresponds to an effective energy derived from demonstration or reward signals, which encourages the model to favor highly preferred behaviors ({\ie}, low-energy states). 
The entropy term $S$ penalizes deviation from the pretrained reference distribution, encouraging the updated distribution to remain broad and close to the base model. 
The temperature $T$ plays a role analogous to the KL-regularization factor $\beta$ in RL, controlling the trade-off between exploiting preferred behaviors and preserving diversity.
Formally, this free-energy objective can be re-expressed as
\begin{equation} \label{eq:free_energy}
\mathcal{F}_x(q)
= \mathbb{E}_{y\sim q(y|x)}[E(x,y)]
+ \beta \, \mathrm{KL}\left[\, q(y|x) \,\|\, p_0(y|x) \right],
\end{equation}
where $\beta$ corresponds to the inverse of temperature $T$ in physics.

The optimal distribution under minimizing \Cref{eq:free_energy} has the form of a Boltzmann-like distribution
\begin{equation} \label{eq:boltzmann_reweighting}
q^*(y|x)
= \frac{1}{Z_E(x)} p_0(y|x) \exp\left[-\frac{1}{\beta}E(x,y)\right].
\end{equation}
A full derivation of \Cref{eq:boltzmann_reweighting} using Lagrange multipliers is provided in Appendix~\ref{sec:appendix}.
Noticeably, the solution in \Cref{eq:boltzmann_reweighting} makes explicit that post-training reweights the reference distribution according to the effective energy $E$. Additionally, it also clearly illustrates that post-training is not selecting behaviors from a uniform space, but using an energy signal to reweight the reference distribution.

\subsection{SFT as A Demonstration-Defined Energy}

For supervised fine-tuning, the target distribution is defined by demonstrations instead of a scalar reward. Let $p_{\rm demo}(y|x)$ denote the empirical or idealized distribution of demonstration responses for prompt $x$. Optimizing SFT objective in \Cref{eq:sft_loss} is equivalent to minimizing
\begin{equation}
\mathbb{E}_{x\sim \mathcal{D}_x} \mathrm{KL} \left[\, p_{\rm demo}(\cdot|x) \,\|\, p_\theta(\cdot|x) \right],
\end{equation}
up to a constant independent of $\theta$.

To connect this statement with the free-energy formulation in \Cref{eq:free_energy}, we ask a reverse question: what effective energy $E_{\rm SFT}(x,y)$ would make the Boltzmann-reweighted solution equal to the demonstration distribution, {\ie}, $q^*_{\mathrm{SFT}}(y|x)=p_{\rm demo}(y|x)$.
This gives us
\begin{equation}
E_{\rm SFT}(x,y) = - \beta \log \frac{p_{\rm demo}(y|x)}{p_0(y|x)} .
\end{equation}
In the above equation, we set the additive constant term to zero as a \emph{convention}, in which the corresponding partition function is exactly $Z_E(x)=1$.
This equivalence should be interpreted carefully. Practical SFT simply minimizes the negative log-likelihood rather than explicitly constructing $E_{\rm SFT}$ to solve a free-energy objective. From this free-energy perspective, SFT can be seen as introducing an effective energy over possible responses. If a response is more likely under the demonstrations than under the base model, it receives lower energy. 

More importantly, this interpretation carries an implicit constraint: whenever $p_{\rm demo}(y|x) > 0$, the reference probability of base model $p_0(y|x)$ must also be nonzero. When the base model assigns a nonzero probability, this ratio is finite and the energy landscape is well-defined, allowing Boltzmann reweighting to smoothly adjust the probability mass across responses. 

However, if this condition is violated -- meaning the demonstrations require behaviors that the base model cannot generate ({\ie}, $p_0(y|x) \to 0$) -- the energy becomes extremely large, creating a singularity. 
In this case, the Boltzmann normalization fails, the local reweighting interpretation breaks down, and post-training can no longer be understood as simply amplifying or suppressing existing behaviors. Thus the singularity naturally marks the boundary between capability elicitation and capability creation. This viewpoint not only clarifies the effective range of the free-energy formulation, but also connects to the concept of accessible support and the categories of regimes developed in \Cref{sec:regimes}.

\subsection{RL as Reward-Defined Energy}

The RL objective in \Cref{eq:rl_loss} is the reward-maximizing counterpart of the free-energy objective in \Cref{eq:free_energy}. Thus, we set the effective energy to the negative reward $E_R(x,y)=-R(x,y)$. The corresponding optimal Boltzmann distribution becomes
\begin{equation} \label{eq:optimal_RLF}
q^*_{\mathrm{RL}}(y|x) = \frac{1}{Z_R(x)} p_0(y|x) \exp\left[\frac{1}{\beta}R(x,y)\right].
\end{equation}
The parameter $\beta$ controls the sharpness of the reward tilt. Large $\beta$ keeps the distribution close to the reference model, while small $\beta$ concentrates probability more aggressively on high-reward outputs.

This shows that, when the $\beta$ is large, {\ie}, KL constraint is strong, RL mainly applies an exponential tilt to the reference distribution according to reward. The higher the reward, the easier a behavior is to amplify; but this amplification remains constrained by the prior mass in $p_0$.

This constraint can be seen by asking when one behavior becomes more likely than another. For two candidate responses $y_a$ and $y_b$, their relative probability under the optimized distribution in \Cref{eq:optimal_RLF} is
\begin{equation} \label{eq:log-odds}
\log \frac{q^*_{\mathrm{RL}}(y_a|x)}{q^*_{\mathrm{RL}}(y_b|x)}
= \log\frac{p_0(y_a|x)}{p_0(y_b|x)}
+ \frac{1}{\beta} \left[ R(x,y_a)-R(x,y_b) \right].
\end{equation}
For a rare behavior to overtake a common behavior, its reward advantage must offset its prior probability disadvantage by $(R(x,y_a)-R(x,y_b)) / \beta$. 
Therefore, reward training can produce seemingly sudden behavioral changes: when the reward gap or temperature parameter $\beta$ crosses a threshold, probability can rapidly move toward behaviors that were previously unlikely under the reference model. 

This is a distributional effect of the exponential tilt. Whether the shifted behavior should be interpreted as elicitation or creation depends on whether that behavior was practically reachable before training, a question we elaborate further in the next section.

\section{Accessible Support and Four Regimes}
\label{sec:regimes}

\subsection{Basins, Tails, and Barriers}

\begin{figure}
    \centering
    \includegraphics[width=0.8\linewidth]{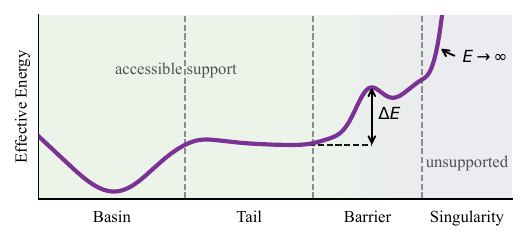}
    \caption{
    Schematic energy landscape for accessible support. 
    Behaviors in basins are easily produced by the base model, while tail behaviors are rare but reachable under larger sampling or search budgets. 
    Barrier regions require crossing low-probability intermediate states. 
    In the unsupported limit, the effective energy becomes divergent, and the local reweighting view no longer applies.}
    \label{fig:demo}
\end{figure}

The free-energy perspective shows that post-training can be understood as reshaping a reference distribution. Crucially, for an autoregressive language model, the softmax distributions assign nonzero probability to many token sequences, so strict mathematical support is too weak for capability analysis. What matters is whether a behavior can be practically produced under finite sampling, decoding, search, optimization, and divergence budgets. 

We therefore use the more practical notion -- accessible support. This is not meant to be a new formal object with a unique estimator. Rather, it is a diagnostic concept, which can be understood as a probability landscape over behavior classes induced by the base model, as illustrated in \Cref{fig:demo}.
\begin{itemize}[leftmargin=*, noitemsep, topsep=0pt]
    \item Some behaviors lie in \textbf{basins}: they have enough probability that the model already produces them stably under ordinary sampling. 
    \item Some behaviors lie in \textbf{tails}: they are rare at pass@1 or greedy decoding, but appear often enough under larger sampling budgets, such as pass@N or best-of-N. 
    \item Other behaviors sit behind \textbf{barriers}: reaching them requires a series of low-probability intermediate steps, long-term planning, or tool invocation. 
\end{itemize}
These regions are not fixed properties of a task alone. They are relative to a base model, a decoding policy, an evaluation signal, and a compute budget. This relativity is why capability claims should specify the mechanism behind the gain rather than only reporting the gain itself.

This landscape suggests four useful regimes for post-training. 
The first two remain within the accessible support inherited from the base model: post-training either stabilizes a high-probability basin or amplifies a low-probability tail. In these cases, the free-energy view remains a useful local description of how training reweights behaviors that the base model can already produce under finite budgets. 
The third regime lies near the boundary of accessible support: the target behavior is not strictly impossible, but it sits behind barriers that ordinary sampling cannot reliably cross. 
The fourth regime is the singular limit, where the target behavior falls outside the support of the base model, the effective energy becomes divergent (a.k.a. singular), and the local reweighting interpretation breaks down.

\subsection{Demonstration-Covered Elicitation}

The first regime occurs when the target behavior already lies in a high-probability basin of the base model and is sufficiently covered by the demonstration or preference signal. 
In this case, post-training does not need to discover a new behavior class; it mainly stabilizes a behavior that the model can already produce reliably. Many forms of instruction following, format control, and stylistic alignment can fall into this regime.

Here, SFT and RL methods may look similar. 
SFT increases the probability of the target behavior by fitting demonstrations. Preference optimization increases it by comparing preferred outputs against alternatives. 
If both demonstrations and rewards point toward the same behavioral basin, then both procedures mainly stabilize the same reachable region. They may still differ in data efficiency, training stability, and optimization risk, but this difference should not be described as a difference between \emph{imitation} and \emph{creation}.

\subsection{Tail Reweighting}

When the target behavior is not sufficiently covered by demonstrations, the second regime still remains within the accessible support of the base model, but in its low-probability tail.
It may be invisible under greedy decoding or pass@1, but appear under repeated sampling, best-of-N, or verifier selection. This regime is especially easy to misread as capability creation, because it can produce large benchmark gains.

The free-energy analysis explains why this improvement can be large. If the target behavior $y_a$ is rare but high-reward, and the competing behavior $y_b$ is common but low-reward, then KL-regularized reward optimization changes their log-odds \eqref{eq:log-odds}. 
Once the reward advantage is large enough to offset the prior probability disadvantage, the rare behavior can quickly become common. Aggregate metrics may jump, but mechanistically this may still be amplification of an existing tail behavior. The key evidence for this regime is whether the base model can already produce the target behavior under a larger sampling budget.

This regime also shows that RL outperforming ordinary SFT does not necessarily mean that RL created a new capability. 
If ordinary SFT demonstrations do not contain rare successes, but a reward model or verifier can select those successes during sampling, then the advantage comes from a tail-selection signal. 
The more appropriate comparison is not ordinary SFT, but SFT on reward-selected, verifier-selected, or search-generated samples \citep{STaR, ToRA}.

\subsection{Barrier-Crossing Discovery}

The third regime concerns behaviors whose reachability depends on how trajectories are generated and preserved. 
Unlike tail reweighting, where a complete target behavior could appear under sufficiently large sampling budgets, barrier-crossing discovery requires the model to maintain a sequence of low-probability or unstable intermediate states before final success is reached. 
Long reasoning, multi-step tool use, interactive problem solving, and code generation that requires exploring partial programs can all have this structure.

This is where the free-energy view reaches the boundary of its local reweighting interpretation. 
A finite energy tilt can amplify candidates that the base model already produces, but it cannot by itself make ordinary sampling reliably preserve the trajectories to reach the target behavior. 
The bottleneck is therefore not only the low probability of the final output, but the reachability of the path that produces it.

Crossing barriers requires changing the trajectory-generation process. 
Search can preserve and expand multiple intermediate candidates; 
process supervision can reward useful steps before the final answer is reached; 
verifiers can filter partial or final solutions; 
tools and environment interaction can introduce information not contained inside the base model. 
RL can be important here, but its role usually comes from being coupled with these mechanisms, not from isolated reward maximization.

This distinction is especially important for reasoning models. 
If the improvement comes from moving already sampled reasoning traces into a high-probability region, it is closer to tail reweighting. 
Otherwise, the improvement is closer to barrier-crossing discovery.

\subsection{Unsupported Regimes}

The fourth regime is the singular limit of the preceding cases. 
Here, the target behaviors fall outside the support of the base model, hence there is no positive probability assigned to it. In the case of $p_0(y|x)=0$, the effective energy becomes divergent and the free-energy reweighting formulation is no longer well-defined.
In this case, post-training lacks the behavioral material needed to operate. Ordinary SFT has no positive examples to imitate, while KL-regularized RL has no candidate trajectories to select and reinforce. 
If neither demonstrations nor generated trajectories reach the relevant region, supervised likelihood fitting has nothing to copy, and sparse reward has nothing to bootstrap.

Unsupported does not mean impossible forever. It only means unreachable relative to the current base model, budget, and training process. 
Stronger pretraining, additional data, different architectures, or other methods can solve this issue to a certain extent. But these changes should be understood as introducing new information or a reachable structure, not as the effect of local post-training alone.

\subsection{Summary}

To sum up, the four regimes form a hierarchy from elicitation to creation: demonstration-covered elicitation stabilizes high-probability basins; tail reweighting amplifies rare but reachable behaviors; barrier-crossing discovery changes the trajectory-generation process; unsupported regimes require new information, tools, or learning processes.

\section{Alternative Views and Counterarguments}

\subsection*{View 1. RL really can create capabilities.}

A strong alternative view is that RL really can create capabilities that SFT struggles to obtain. In mathematical reasoning, code generation, and many other scenarios, it is easier to provide correctness feedback than to write optimal demonstrations \citep{DeepSeekMath, codex}. As long as there is a verifier or environment reward, RL may discover strategies that humans did not demonstrate.\looseness=-1

We agree that RL can participate in capability creation, but we disagree with attributing creation to the label \emph{RL} itself. The key question is whether reward is coupled with mechanisms that change reachability.
In such cases, RL is one part of a larger system: reward signals, jointly with search, interaction, or new information, change the behavior-generation process.

Therefore, strong cases of RL-driven capability creation are better understood as barrier-crossing discovery rather than local reward reweighting. 
Reward matters, but it works because the system can expose, select, and reinforce trajectories that ordinary sampling previously struggled to reach \citep{trinh2024solving, ARGS}. This interpretation preserves the advantage of RL while avoiding the claim that all post-RL performance gains are capability creation.

\subsection*{View 2. SFT is only imitation and therefore inherently weaker.}

Another view holds that SFT maximizes demonstration likelihood and therefore can only copy behaviors already contained in the dataset; it cannot go beyond demonstrations \citep{LIMA}. 
This view is correct when demonstrations are narrow or low-quality. If the target behavior does not appear in the demonstrations and is not exposed by the data-generation process, ordinary SFT has no direct mechanism for discovering it.

But this does not show that SFT is inherently weak. SFT imitates a distribution, but the supervised loss itself does not tell us where the data came from. The same supervised objective can be applied to human-written answers, search-generated trajectories, or reward-filtered samples, but these data sources carry different information and induce different reachability structures \citep{STaR, LIMO, quan2024automatically}.

Thus, comparisons between SFT and RL must specify the source of demonstrations. If weak-demonstration SFT is compared against RL with extra features, the experiment may conflate the effect of reward optimization with the effect of additional candidates. 
The framework of our paper requires these factors to be separated.

\subsection*{View 3. The physics language is only metaphor.}

One might worry that terms such as energy, entropy, basins, and barriers are too loose. Language models are not thermodynamic systems and are not in physical equilibrium. If these terms are only rhetorical, they do not improve the rigor of the argument.

This concern is reasonable. This paper uses physics language only in a restricted sense: the objective
\begin{equation}
\mathbb{E}_{q}[E(x,y)] + \beta \, \mathrm{KL}[\, q(y|x) \,\|\, p_0(y|x)]
\end{equation}
does have a free-energy-like structure, and its optimum is a Boltzmann reweighting of the reference distribution. 
Thus, \emph{basins}, \emph{tails}, and \emph{barriers} are not literal thermodynamic concepts, but shorthand for describing distributional structure and finite-budget reachability. Their role is to help distinguish local reweighting, mode switching, and support expansion.

We do not assume that training reaches thermodynamic equilibrium, nor that all performance jumps are genuine phase transitions. Free-energy language is only an organizing principle. Phenomena that this form cannot explain, especially reachability changes from new information, must be explained by additional mechanisms.

\subsection*{View 4. Optimization dynamics are underestimated.}

A final objection is that SFT and RL have very different optimization dynamics, which can strongly affect practical outcomes. We agree that optimization dynamics matter. We do not replace optimization analysis with a distributional view. Instead, it separates two levels: how an algorithm moves through parameter space, and how it ultimately changes the behavioral distribution and accessible support.\looseness=-1

Therefore, optimization dynamics should be interpreted relative to the regime. 
In basin regimes, many optimization paths may stabilize the same behavior. 
In tail regimes, optimization succeeds only if rare successes are sampled and selected. 
In barrier-crossing regimes, optimization must be coupled with a process that changes the trajectory distribution. 
In unsupported regimes, no optimizer can amplify behaviors that are never generated or evaluated.

These views do not overturn the paper’s position, but delimit its scope. 
Our claim is narrower: claims about capability creation need to specify their mechanism. A performance improvement may come from demonstration-covered stabilization, tail amplification, barrier-crossing discovery, or new information and tools. Different explanations require different evidence.

\section{Conclusion}

The debate over SFT and RL in post-training is often framed as a debate between imitation and discovery. We have argued that this framing obscures the more important distinction between capability elicitation and creation. 
SFT and RL are algorithmic categories; elicitation and creation are claims about how a post-training procedure transforms the model's behavioral distribution and accessible support. 
A method should not be credited with creating a capability merely because it improves performance after reward optimization. Likewise, SFT should not be dismissed as weak imitation without examining the source and coverage of its demonstrations.

The free-energy perspective provides a compact way to make this distinction precise. 
Both demonstration-driven fitting and KL-regularized reward optimization can be viewed as transformations of a pretrained reference distribution under an external preference signal. 
SFT specifies low-energy behavior through demonstrations; RL specifies low-energy behavior through reward. When the resulting update remains close to the reference distribution, its primary effect is to reweight probability mass within the base model's accessible support.

This view leads to a regime-based understanding of post-training. 
In demonstration-covered regimes, SFT and RL may behave similarly because both stabilize behaviors that are already accessible. 
In tail-reweighting regimes, reward-based methods can outperform ordinary SFT by amplifying rare but already reachable behaviors. 
In barrier-crossing regimes, capability gains require mechanisms that alter the trajectory-generation process itself. 
In unsupported regimes, neither ordinary SFT nor RL should be expected to create a capability without additional information or changes to the learning process.\looseness=-1

The central question for post-training is therefore not whether RL is categorically more powerful than SFT. It is whether a method merely changes the probabilities of behaviors that were already reachable, or changes what the model can reach. 
Future work on post-training should make this distinction explicit. Performance improvement is evidence that post-training changed what the model tends to do. 
But performance improvement alone is not sufficient evidence that post-training changed what the model can practically reach. Capability creation is a stronger claim than performance improvement.

\clearpage
\renewcommand*{\bibfont}{\small}
\printbibliography

\clearpage
\appendix
\section{Detailed Derivation for the Boltzmann Reweighting Solution}
\label{sec:appendix}

To derive the minimizer of free-energy, we introduce a Lagrange multiplier $\lambda$ for the normalization constraint $\sum_y q(y|x)=1$ and minimize the free energy in \Cref{eq:free_energy} as: 
\begin{equation}
\mathcal{L}_x(q,\lambda)
= \sum_y q(y|x)E(x,y)
+ \beta \sum_y q(y|x)\log\frac{q(y|x)}{p_0(y|x)}
+ \lambda\left[\sum_y q(y|x)-1\right].
\end{equation}
Taking the functional derivative with respect to $q(y|x)$ and setting it to zero gives
\begin{equation}
E(x,y) + \beta \left[ \log\frac{q(y|x)}{p_0(y|x)} +1 \right] + \lambda =0.
\end{equation}
Solving for $q(y|x)$ gives
\begin{equation}
q(y|x) = p_0(y|x) \exp \left[ -\frac{E(x,y)}{\beta} -1 -\frac{\lambda}{\beta} \right].
\end{equation}
Since the factor $\exp[-1-\lambda/\beta]$ is independent of $y$, it only affects the overall normalization. 
We therefore absorb it into a normalization constant and write the optimal distribution as
\begin{equation}
q^*(y|x)
= \frac{1}{Z_E(x)} p_0(y|x) \exp\left[-\frac{1}{\beta}E(x,y)\right],
\end{equation}
where the partition function is defined by the normalization condition as
\begin{equation}
Z_E(x) = \sum_y p_0(y|x) \exp\left[-\frac{1}{\beta}E(x,y)\right].
\end{equation}

This expression is the formal basis for the local reweighting interpretation. Outputs with high prior under $p_0$ and low energy under $E$ become more likely. Conversely, outputs with negligible prior remain hard to make visible unless the energy tilt is sufficiently strong.

\end{document}